\def\year{2018}\relax
\documentclass[letterpaper]{article} 
\usepackage{aaai18}  
\usepackage{times}  
\usepackage{helvet}  
\usepackage{courier}  
\usepackage{url}  
\usepackage{graphicx}  
\usepackage{amsmath}
\usepackage{graphicx}
\usepackage[colorinlistoftodos]{todonotes}
\usepackage[]{algorithm2e}
\usepackage{array}
\newcolumntype{L}[1]{>{\raggedright\let\newline\\\arraybackslash\hspace{0pt}}m{#1}}
\newcolumntype{C}[1]{>{\centering\let\newline\\\arraybackslash\hspace{0pt}}m{#1}}
\newcolumntype{R}[1]{>{\raggedleft\let\newline\\\arraybackslash\hspace{0pt}}m{#1}}
\usepackage{comment}

\newif\ifarxiv
\arxivfalse

\newif\ifsvm
\svmtrue

\begin{document}
%
\title{StarSpace:\\ Embed All The Things!}
\author{Ledell Wu, Adam Fisch, Sumit Chopra, Keith Adams,
Antoine Bordes and Jason Weston\\
Facebook AI Research
}

\maketitle
\begin{abstract}
We present StarSpace, a general-purpose neural embedding model that can solve a wide variety of problems: 
labeling tasks such as text classification,
ranking tasks such as information retrieval/web search,
collaborative filtering-based  or content-based recommendation,
embedding of multi-relational graphs, 
and learning word, sentence or document level embeddings.
In each case the model works by embedding those entities comprised of discrete features and comparing them against each other 
-- learning similarities dependent on the task.
Empirical results on a number of tasks show that StarSpace is highly competitive with existing methods, whilst also being generally applicable to new cases where those methods are not.
\end{abstract}

\section{Introduction}
\label{sec:intro}

We introduce StarSpace, a neural embedding model that is general enough to solve a wide variety of problems:
\begin{itemize}
\item Text classification, or other labeling tasks, e.g. sentiment classification.
\item Ranking of sets of entities, e.g. ranking web documents given a query.
\item Collaborative filtering-based recommendation, e.g. recommending documents, music or videos.
\item Content-based recommendation where content is defined with discrete features, e.g. words of documents.
\item Embedding graphs, e.g. multi-relational graphs such as Freebase.
\item Learning word, sentence or document embeddings.
\end{itemize}

StarSpace can be viewed as a straight-forward and  
efficient strong baseline for any of these tasks. In experiments it is shown to be on par with or outperforming several competing methods, whilst being generally applicable to cases where many of those methods are not.

The method works by learning entity embeddings with discrete feature representations from relations among collections of those entities directly for the task of ranking or classification of interest.
In the general case, StarSpace embeds entities of different types into a vectorial embedding space, hence the ``star'' (``*'', meaning all types) and ``space'' in the name, and in that common space compares them against each other. It learns to rank a set of entities, documents or objects given a query entity, document or object, where the query is not necessarily of the same type as the items in the set.

We evaluate the quality of our approach on six
different tasks, namely text classification, link prediction in knowledge bases, document recommendation, 
article search, sentence matching
and learning general sentence embeddings. StarSpace is available as an open-source project at 
\url{https://github.com/facebookresearch/Starspace}.

\section{Related Work}
\label{sec:related}

Latent text representations, or \textit{embeddings}, are vectorial representations of words or documents, traditionally learned in an unsupervised way over large corpora. Work on neural embeddings in this domain includes \cite{bengio2003}, \cite{collobert2011}, word2vec \cite{word2vec} and more recently fastText \cite{fasttext-unsup}. 
In our experiments we compare to word2vec 
and fastText 
as representative scalable models for unsupervised embeddings; we also compare on the SentEval tasks \cite{infersent} against a wide range of unsupervised models for sentence embedding.

In the domain of supervised embeddings, 
SSI \cite{bai2009supervised} and WSABIE \cite{wsabie} are early approaches that
showed promise in NLP and information retrieval tasks (\cite{weston2013connecting}, \cite{hermann2014}). Several more recent works including \cite{tang2015document}, 
 \cite{zhang2015text}, \cite{conneau2016very},
TagSpace \cite{tagspace} 
and fastText \cite{fasttext} have yielded good results on classification tasks such as sentiment analysis or hashtag prediction.

In the domain of recommendation, embedding models have had a large degree of success, starting from 
SVD \cite{goldberg2001eigentaste} and its improvements such as SVD++ \cite{koren2015advances}, as well as a host of other techniques, e.g.
\cite{rendle2010factorization,lawrence2009non,shi2012climf}.
Many of those methods have focused on the collaborative filtering setup where user IDs and movie IDs have individual embeddings,
such as in the Netflix challenge setup
(see e.g., \cite{koren2015advances}, 
and so new users or items cannot naturally be incorporated.
We show how StarSpace  can naturally cater for both that setting and the content-based setting where users and items are represented as features, and hence have natural out-of-sample extensions rather than considering only a fixed set.

Performing link prediction in knowledge bases (KBs) with 
embedding-based methods 
has also shown promising results in recent years. A series of work has been done in this direction, such as \cite{transE} and \cite{garcia2015composing}. In our work, we show that StarSpace  
can be used for this task as well, 
outperforming several methods, and matching
the TransE method presented in \cite{transE}.

\section{Model}
\label{sec:model}

The StarSpace model consists of 
learning {\em entities}, each of which is described
by a set of discrete {\em features} (bag-of-features) coming from a fixed-length dictionary. 
An entity such as a document or a sentence can be described by a bag of words or $n$-grams, an entity such as a user can be described by the bag of documents, movies or items they have liked, and so forth. 
Importantly, the StarSpace model is free to compare entities of {\em different kinds}.
For example, a user entity can be compared with an item entity (recommendation), or a document entity with label entities (text classification), and so on. This is done by learning to embed them in the same space such that comparisons are meaningful -- by optimizing with respect to
the metric of interest.

Denoting the dictionary of ${\cal D}$ features as $F$ which is a ${\cal D} \times d$ matrix, where $F_i$ indexes the $i^{th}$ feature (row), yielding its $d$-dimensional embedding, we embed an entity $a$ 
  with $\sum_{i \in a} F_i$. 

That is, like other embedding models, our model starts by assigning a $d$-dimensional vector to each of the discrete features in the set that we want to embed directly (which we call a \textit{dictionary}, it can contain features like words, etc.). Entities comprised of features (such as documents) 
are represented by a bag-of-features of the features in the dictionary and their embeddings are learned implicitly.  Note an entity could consist of a single (unique) feature like a single word, name or user or item ID if desired.

To train our model, we need to learn to {\em compare} entities. Specifically, we want to minimize the following loss function:
\[
\label{math:loss}
\sum_{\substack{(a,b) \in E^+\\ b^{-} \in E^{-}}} L^{batch}(sim(a, b), sim(a, b^{-}_1), \dots, sim(a, b^{-}_k))
\]

There are several ingredients to this recipe:
\begin{itemize}
\item The generator of positive entity pairs $(a,b)$ coming from the set ${E^+}$. This is task dependent and will be described subsequently.
\item The generator of negative entities $b^-_i$ coming from the set ${E^-}$.
We utilize a $k$-negative sampling strategy \cite{word2vec} to select $k$ such negative pairs for each batch update. We select randomly from within the set of entities that can appear in the second argument of the similarity function (e.g., for text labeling tasks $a$ are documents and $b$ are labels, so we sample $b^{-}$ from the set of labels). An analysis of the impact of $k$ is given in Sec. \ref{sec:experiment}.

\item The similarity function $sim(\cdot, \cdot)$. In our system, we have implemented both cosine similarity and inner product, and selected the choice as a hyperparameter. Generally, they work similarly well for small numbers of label features (e.g. for classification), while cosine works better for larger numbers, e.g. for sentence or document similarity.

\item The loss function $L_{batch}$ that compares the positive pair $(a,b)$ with the negative pairs $(a, b^{-}_i)$, $i=1,\dots, k$. We also implement two possibilities: margin ranking loss (i.e. $\max(0,\mu - sim(a,b)$, where $\mu$ is the margin parameter), and negative log loss of softmax. All experiments use the former as it performed on par or better.

\end{itemize}

We optimize by stochastic gradient descent (SGD), i.e., each SGD step is one sample from $E^{+}$ in the outer sum, using Adagrad \cite{adagrad} and hogwild \cite{hogwild} over multiple CPUs. We also apply a max norm of the embeddings to restrict the vectors learned to lie in a ball of radius $r$ in space $R^d$, as in other works, e.g. \cite{wsabie}.

At test time, one can use the learned function $sim(\cdot,\cdot)$ to measure similarity between entities.
For example, for classification, a label is predicted at test time for a given input $a$ using $\max_{\hat{b}} sim(a, {\hat{b}})$ over the set of possible labels $\hat{b}$. Or in general, for ranking one can sort entities by their similarity. 
Alternatively the embedding vectors can be used directly for some other downstream task, e.g., as is typically done with word embedding models. However, if $sim(\cdot,\cdot)$ directly fits the needs of your application, this is recommended as this is the objective that StarSpace is trained to be good at.

We now describe how this model can be applied to a wide variety of tasks, in each case describing how the generators E$^+$ and E$^-$ work for that setting.

\paragraph{Multiclass Classification (e.g. Text Classification)} The positive pair generator comes directly from a training set of labeled data specifying $(a,b)$ pairs where $a$ are documents (bags-of-words) and $b$ are labels (singleton features). Negative entities $b^-$ are sampled from the set of possible labels.

\paragraph{Multilabel Classification} 
In this case, each document $a$ can have multiple positive labels,
one of them is sampled as $b$ at each SGD step
to implement multilabel classification.

\paragraph{Collaborative Filtering-based Recommendation}
The training data consists of a set of users, where each user is described by a bag of items (described as unique features from the dictionary) that the user likes.
The positive pair generator picks a user, selects $a$ to be the unique singleton feature for that user ID, and a single item that they like as $b$. Negative entities $b^-$ are sampled from the set of possible items.

\paragraph{Collaborative Filtering-based Recommendation with out-of-sample user extension}
One problem with classical collaborative filtering is that it does not generalize to new users, as a separate embedding is learned for each user ID.
Using the same training data as before, one can learn an alternative model using StarSpace.
The positive pair generator instead picks a user, selects $a$ as all the items they like except one, and $b$ as the left out item. That is, the model learns to estimate if a user would like an item by modeling the user not as a single embedding based on their ID, but by representing the user as the sum of embeddings of items they like.

\paragraph{Content-based Recommendation} This task consists of a set of users, where each user is described by a bag of items, where each item is described by a bag of features from the dictionary (rather than being a unique feature). 
For example, for document recommendation, each user is described by the bag-of-documents they like, while each document is described by the bag-of-words it contains.
Now $a$ can be selected as all of the items except one, and $b$ as the left out item. The system now extends to both new items and new users as both are featurized.

\paragraph{Multi-Relational Knowledge Graphs (e.g. Link Prediction)} Given a graph of $(h,r,t)$ triples, consisting of a head concept $h$, a relation $r$ and a tail concept $t$, e.g. {\em (Beyonc\'e, born-in, Houston)}, one can learn embeddings of that graph.
Instantiations of $h$, $r$ and $t$ are all defined as unique features in the dictionary. We select uniformly at random either: (i) 
$a$ consists of the bag of features $h$ and $r$, while $b$ consists only of $t$; or (ii) $a$ consists of $h$, and $b$ consists of $r$ and $t$. Negative entities $b^-$ are sampled from the set of possible concepts. The learnt embeddings can then be used to answer link prediction questions such as {\em (Beyonc\'e, born-in, ?)} or {\em (?, born-in, Houston)} via the learnt  function $sim(a, b)$.

\paragraph{Information Retrieval (e.g. Document Search) and Document Embeddings}
Given supervised training data consisting of (search keywords, relevant document) pairs one can directly train an information retrieval model: $a$ contains the search keywords, $b$ is a relevant document and $b^{-}$ are other irrelevant documents. If only unsupervised training data is available consisting of a set of unlabeled documents, an alternative is to select $a$ as random keywords from the document and $b$ as the remaining words.
Note that both these approaches implicitly learn document embeddings which could be used for other purposes.

\paragraph{Learning Word Embeddings}
We can also use StarSpace to learn unsupervised word embeddings using training data consisting of raw text.
We select $a$ as a window of words (e.g., four words,  two either side of a middle word), and $b$ as the middle word, following  \cite{collobert2011,word2vec,fasttext-unsup}.

\paragraph{Learning Sentence Embeddings}
Learning word embeddings (e.g. as above) and using them to embed sentences does not seem optimal when you can learn sentence embeddings directly. Given a training set of unlabeled documents, each consisting of sentences, we select $a$ and $b$ as a pair of sentences both coming from the same document; $b^{-}$ are sentences coming from other documents. The intuition is that semantic similarity between sentences is shared within a document (one can also only select sentences within a certain distance of each other if documents are very long). Further, 
the embeddings will automatically be optimized for sets of words of sentence length, so train time matches test time, rather than training with short windows as typically learned with word embeddings -- window-based embeddings can deteriorate when the sum of words in a sentence gets too large.

\paragraph{Multi-Task Learning}
Any of these tasks can be combined, and trained at the same time if they share some features in the base dictionary $F$. For example one could combine supervised classification with unsupervised word or sentence embedding, to give semi-supervised learning.

\section{Experiments}
\label{sec:experiment}

\subsection{Text Classification}

We employ StarSpace for the task of text classification and compare it with a host of competing methods, including fastText, on three datasets which were all previously used in \cite{fasttext}. 
To ensure fair comparison, we use an identical dictionary  to fastText and use the same implementation of $n$-grams and pruning (those features are implemented in our open-source distribution of StarSpace).
In these experiments we set the dimension of embeddings to be 10, as in \cite{fasttext}.

We use three datasets:
\begin{itemize}
\item AG news\footnote{\tiny\url{http://www.di.unipi.it/˜gulli/AG_corpus_of_news_articles.html}} is a 4 class text classification task given title and description fields as input.
It consists of 120K training examples, 7600 test examples, 4 classes, $\sim$100K words and 5M tokens in total.
\item DBpedia \cite{lehmann2015dbpedia} is a 14 class classification problem given the title and abstract of Wikipedia articles as input. It consists of 560K training examples, 70k test examples, 14 classes, $\sim$800K words and 32M tokens in total.
\item The Yelp reviews dataset is obtained from the 2015 Yelp Dataset Challenge\footnote{\tiny\url{https://www.yelp.com/dataset_challenge}}.
The task is to  predict the full number of stars the user has given (from 1 to 5).  It consists of 1.2M training examples, 157k test examples,  5 classes, $\sim$500K words and 193M tokens in total.
\end{itemize}

Results are given in Table \ref{tab:classification-acc}.
Baselines are quoted from the literature (some methods are only reported on AG news and DBPedia, others only on Yelp15). 
StarSpace outperforms a number of methods, and performs similarly to fastText. We measure the training speed 
for $n$-grams $> 1$ in Table \ref{tab:classification-time}. 
fastText and StarSpace are both efficient compared to deep learning approaches, e.g.
\cite{zhang2015text}
 takes 5h per epoch on DBpedia,  375x slower than StarSpace.
Still, fastText is faster than StarSpace.
However, as we will see in the following sections, StarSpace is a more general system.

\begin{table*}[t!]
\centering
\begin{tabular}{|l|r|r|r|r|r|}
\hline
 Metric & Hits@1 &	Hits@10 &	Hits@20 &	Mean Rank & Training Time\\\hline
{\it Unsupervised methods} & & & & & \\
TFIDF & 0.97\% & 3.3\% & 4.3\% & 3921.9 & - \\
word2vec & 0.5\% &	1.2\% &	1.7\% &	4161.3 &  - \\
fastText (public Wikipedia model) &	0.5\% &	1.7\% &	2.5\% & 4154.4 & - \\
fastText (our dataset) &	0.79\% &	2.5\% &	3.7\% &	3910.9 & 4h30m \\
Tagspace${\dagger}$	& 1.1\% &	2.7\% &	4.1\% &	3455.6 & -\\
\hline
{\it Supervised methods} & & & & & \\
SVM Ranker: BoW features & 0.99\% & 3.3\% & 4.6\% & 2440.1 & - \\
\ifsvm
SVM Ranker: fastText features (our dataset) & 0.92\% & 3.3\% & 4.2\% & 3833.8 
& - \\
\fi
StarSpace &	3.1\%	& 12.6\%	& 17.6\% &	1704.2 & 12h18m \\
\hline
\end{tabular}
\caption{\label{tab:recommendation}Test metrics and training time on 
the Content-based Document Recommendation task. ${\dagger}$ Tagspace training is supervised but for another task (hashtag prediction) not our task of interest here.
\label{tab:feedspace}
}

\end{table*}
\begin{table}[h]
\centering
\begin{tabular}{|l|r|r|r|}
\hline
Model	& AG news	& DBpedia	& Yelp15 \\\hline
BoW*  & 88.8 & 96.6 & - \\
ngrams* & 92.0 & 98.6 & - \\
ngrams TFIDF*  & 92.4 & 98.7 & - \\
char-CNN* & 87.2 & 98.3 & - \\
char-CRNN$\star$  & 91.4 & 98.6 & - \\
VDCNN$\diamond$ & 91.3 & 98.7 & - \\\hline
SVM+TF$\dagger$  & - & - & 62.4 \\
CNN$\dagger$  & - & - & 61.5 \\
Conv-GRNN$\dagger$  & - & - & 66.0 \\
LSTM-GRNN$\dagger$ & - & - & 67.6 \\\hline
fastText (ngrams=1)$\ddagger$ &	91.5	& 98.1 & $^{**}$62.2 \\
StarSpace (ngrams=1) & 91.6 &	98.3 &	62.4 \\
fastText (ngrams=2)$\ddagger$	& 92.5 &	98.6 & - \\
StarSpace (ngrams=2) &	92.7 &	98.6 & - \\
fastText (ngrams=5)$\ddagger$	& - &  - &		66.6 \\
StarSpace (ngrams=5) &  - & -	&  65.3 \\\hline
\end{tabular}
\caption{\label{tab:classification-acc}Text classification test accuracy. * indicates models from \cite{zhang2015text}; $\star$  from \cite{xiao2016efficient}; $\diamond$  from \cite{conneau2016very}; $\dagger$ from \cite{tang2015document}; $\ddagger$ from \cite{fasttext}; $^{**}$ we ran ourselves.}
\end{table}

\begin{table}[h]
\centering
\begin{tabular}{|l|r|r|r|}
\hline
Training time	& ag news	& dbpedia	& Yelp15 \\\hline
fastText (ngrams=2)	& 2s &	10s & \\
StarSpace (ngrams=2) &	4s & 34s & \\
fastText (ngrams=5)	& & &	2m01s \\
StarSpace (ngrams=5)& &	&   3m38s \\\hline

\end{tabular}
\caption{\label{tab:classification-time}Training speed on the text classification tasks.}
\end{table}

\subsection{Content-based Document Recommendation}

We consider the task of recommending new documents to a user 
given their past history of liked documents.
We follow a very similar process described in \cite{tagspace} in our experiment. 
The data for this task is comprised of anonymized two-weeks long interaction histories for a subset of people on a popular social networking service. For each of the 641,385 people considered, we collected the text of public articles that s/he clicked to read, giving a total of 3,119,909 articles. Given the person's trailing $(n-1)$ clicked articles, we use our model to predict the $n$'th article by ranking it against 10,000 other unrelated articles, and evaluate using ranking metrics. The score of the $n$'th article is obtained by applying StarSpace: the input $a$ is the previous $(n-1)$ articles, and the output $b$ is the $n$'th candidate article.  We measure the results by computing hits@k,
i.e. the proportion of correct entities ranked in the top k for $k=$ 1, 10, 20,
and the mean predicted rank of the clicked article among the 10,000 articles.

As this is not a classification task (i.e. there are not a fixed set of labels to classify amongst, but a variable set of never seen before documents to rank per user) we cannot use supervised classification models directly. Starspace however can deal directly with this task, which is one of its major benefits.
Following \cite{tagspace}, we hence use the following models as baselines:
\begin{itemize}
\item Word2vec model. 
We use the publicly available word2vec model trained on Google News articles\footnote{\tiny\url{https://code.google.com/archive/p/word2vec/}}, and use the word embeddings to generate article embeddings (by bag-of-words) and users' embedding (by bag-of-articles in users' click history). We then use cosine similarity for ranking.
\item Unsupervised fastText model. 
We try both the previously trained publicly available model on Wikipedia\footnote{\tiny\url{https://github.com/facebookresearch/fastText/blob/master/pretrained-vectors.md}}, and train on our own dataset. Unsupervised fastText is an enhancement of word2Vec that also includes subwords. 
\ifsvm
\item Linear SVM ranker, using either bag-of-words features or fastText embeddings (component-wise multiplication of $a$'s and $b$'s features, which are of the same dimension).  
\else
\item Linear SVM ranker using bag-of-words features (component-wise multiplication of $a$'s and $b$'s features, which are of the same dimension).  
\fi

\item Tagspace model trained on a hashtag task, and then the embeddings are used for document recommendation, a reproduction of the setup in \cite{tagspace}. In that work, the Tagspace model was shown to outperform word2vec.
\item TFIDF bag-of-words cosine similarity model.
\end{itemize}

\begin{table*}[t]
\centering
\begin{tabular}{|l|r|r|r|r|r|}
\hline
Metric & Hits@10 r.	& Mean Rank r. &  Hits@10 f. & Mean Rank f. & Train Time\\\hline
SE* \cite{bordes2011learning} &  28.8\% & 273 & 39.8\% & 162 & - \\
SME(LINEAR)* \cite{bordes2014semantic} & 30.7\% & 274 & 40.8\% & 154 & -\\
SME(BILINEAR)* \cite{bordes2014semantic} & 31.3\% & 284 & 41.3\% & 158 & - \\
LFM* \cite{jenatton2012latent} & 26.0\% & 283 & 33.1\% & 164 & -\\
RESCAL$\dagger$ \cite{rescal} & - & - & 58.7\% & - & 
\\\hline
TransE (dim=50) & 47.4\% & 212.4 & 71.8\% & 63.9 & 1m27m \\
TransE (dim=100) & 51.1\% & 225.2 & 82.8\% & 72.2 & 1h44m \\
TransE (dim=200) & 51.2\% & 234.3 & 83.2\% & 75.6 & 2h50m \\
StarSpace (dim=50) & 45.7\% & 191.2 & 74.2\% & 70.0 & 1h21m \\
StarSpace (dim=100) & 50.8\% & 209.5 & 83.8\% & 62.9 & 2h35m \\
StarSpace (dim=200) & 52.1\% & 245.8 & 83.0\% & 62.1 & 2h41m
\\\hline 
\end{tabular}
\caption{\label{tab:fb15k}Test metrics on Freebase 15k dataset. * indicates results cited from \cite{transE}. $\dagger$ indicates results cited from \cite{hole}.
\label{tab:fb-result}
}
\end{table*}

\begin{table*}[t!]
\centering
\begin{tabular}{|l|r|r|r|r|r|r|r|r|r|}
\hline
 K & 1 & 5 & 10 & 25 & 50 & 100 & 250 & 500 & 1000 \\
\hline
Epochs & 3260 & 711 & 318 & 130 & 69 & 34 & 13 & 7 & 4 \\
 hit@10 & 67.05\% & 68.08\% & 68.13\% & 67.63\% & 69.05\% & 66.99\% & 63.95\% & 60.32\% & 54.14\% \\
\hline
\end{tabular}
\caption{Adapting the number of negative samples $k$ for a 50-dim model for 1 hour of training on Freebase 15k.}
\label{tab:complexity}
\end{table*}

\begin{table*}[h]
\centering
\begin{tabular}{|l|r|r|r|r|r|}
\hline
 Metric & Hits@1 &	Hits@10 &	Hits@20 &	Mean Rank & Training Time\\\hline 
{\it Unsupervised methods} & & & & & \\
TFIDF & 56.63\% & 72.80\% & 76.16\% & 578.98 & -{\tiny } \\
fastText (public Wikipedia model) & 18.08\% & 36.36\% & 42.97\% & 987.27 & -{\tiny  } \\
fastText (our dataset) & 16.89\% & 37.60\% & 45.25\% & 786.77 &  ~40h\\
\hline
{\it Supervised method} & & & & & \\
SVM Ranker BoW features & 56.73\% & 69.24\% & 71.86\% & 723.47 & - \\
SVM Ranker: fastText features (public) & 18.44\% & 37.80\% & 45.91\% & 887.96 & - \\ 
StarSpace & 56.75\% & 78.14\% & 83.15\% & 122.26 & ~89h\\
\hline
\end{tabular}
\caption{Test metrics and training time on Wikipedia Article Search (Task 1).
\label{tab:wiki1}
}
\end{table*}

\begin{table*}[h]
\centering
\begin{tabular}{|l|r|r|r|r|r|}
\hline
 Metric & Hits@1 &	Hits@10 &	Hits@20 &	Mean Rank & Training Time\\\hline 
{\it Unsupervised methods} & & & & & \\
TFIDF & 24.79\% & 35.53\% & 38.25\% & 2523.68 & -\\
fastText (public Wikipedia model) & 5.77\% & 14.08\% & 17.79\% & 2393.38 &  -\\
fastText (our dataset) & 5.47\% & 13.54\% & 17.60\% & 2363.74  &  ~40h\\
StarSpace (word-level training) & 5.89\% & 16.41\% & 20.60\% & 1614.21 & ~45h \\
\hline
{\it Supervised methods} & & & & & \\
SVM Ranker BoW features & 26.36\% & 36.48\% & 39.25\%& 2368.37 & - \\
SVM Ranker: fastText features (public) & 5.81\% & 12.14\% & 15.20\% & 1442.05 & - \\
StarSpace (sentence pair training) & 30.07\% & 50.89\% & 57.60\% & 422.00 & ~36h \\
StarSpace (word+sentence training) & 25.54\% & 45.21\%& 52.08\% & 484.27 & ~69h\\
\hline
\end{tabular}
\caption{Test metrics and training time on Wikipedia Sentence Matching (Task 2).
\label{tab:wiki2}
}
\end{table*}

For fair comparison, we set the dimension of all embedding models to be 300.
We show the results of our StarSpace 
model comparing with the baseline models in Table \ref{tab:feedspace}. Training time for StarSpace and fastText \cite{fasttext-unsup} trained on our dataset is also provided. 

Tagspace was previously shown to
provide superior performance to word2vec, and we observe the same result here. Unsupervised FastText, which is an enhancement of word2vec is also slightly inferior to Tagspace, but better than word2vec. However, StarSpace, which is naturally more suited to this task, outperforms all those methods, including Tagspace and SVMs by a significant margin. Overall, from the evaluation one can see that unsupervised methods of learning word embeddings are inferior to training specifically for the document recommendation task at hand, which StarSpace does.

\subsection{Link Prediction: Embedding Multi-relation Knowledge Graphs}
We show that one can also use StarSpace on tasks of knowledge representation. We use the Freebase 15k dataset from \cite{transE}, which consists of a collection of triplets (head, relation\_type, tail) extracted from Freebase\footnote{\url{http://www.freebase.com}}. This data set can be seen as a 3-mode tensor depicting ternary relationships between synsets.
There are 14,951 concepts (mids) and 1,345 relation types among them. The training set contains 483,142 triplets, the validation set 50,000 and the test set 59,071. As described in \cite{transE}, 
evaluation is performed by, for each test triplet, removing the head and replacing by each of the entities in the dictionary in turn. Scores for those corrupted triplets are first computed by the models and then sorted; the rank of the correct entity is finally stored. This whole procedure is repeated while removing the tail instead of the head. We report the mean of those predicted ranks and the hits@10. 
We also conduct a filtered evaluation that is the same, except all other valid heads or tails from the train or test set are discarded in the ranking, following \cite{transE}.

We compare with a number of methods, including transE presented in \cite{transE}.
TransE was shown to outperform  RESCAL \cite{rescal}, RFM \cite{jenatton2012latent}, SE \cite{bordes2011learning} and SME \cite{bordes2014semantic} and is considered a standard benchmark method. TransE uses an L2 similarity $||$head + relation - tail$||^2$ and SGD updates with single entity corruptions of head or tail that should have a larger distance. In contrast, StarSpace uses a dot product, $k$-negative sampling, and two different embeddings to represent the relation entity, depending on whether it appears in $a$ or $b$.

The results are given in Table \ref{tab:fb-result}.
Results for 
SE, SME and LFM are reported from \cite{transE} and optimize
the dimension from the choices 20, 50 and 75 as a hyperparameter.
RESCAL is reported from \cite{hole}. For TransE we ran it ourselves so that we could report the results for different embedding dimensions, and because we obtained better results by fine tuning it than previously reported. 
Comparing TransE and StarSpace for the same embedding dimension, these two methods then give similar performance. 
Note there are some recent improved results on this dataset using larger embeddings 
\cite{kadlec2017knowledge} or more complex, but less general, methods \cite{shen2017modeling}.

\paragraph{Influence of $k$}

In this section, we ran experiments on the Freebase 15k dataset to illustrate the complexity of our model in terms of the number of negative search examples. We set $dim = 50$, and the max training time of the algorithm to be 1 hour for all experients. We report the number of epochs the algorithm completes within the time limit and the best filtered hits@10 result over possible learning rate choices, for different $k$ (number of negatives searched for each positive training example). We set $k = [1, 5, 10, 25, 50, 100, 250, 500, 1000]$. 

The result is presented in Table \ref{tab:complexity}. We observe that the number of epochs finished within the 1 hour training time constraint is close to an inverse linear function of $k$. In this particular setup, [1, 100] is a good range of $k$ and the best result is achieved at $K=50$.

\begin{table*}[ht]
\scriptsize
\begin{tabular}{L{5.5cm}|L{5.5cm}|L{5.5cm}}
\hline
Input Query & StarSpace result & fastText result  \\\hline
She is the 1962 Blue Swords champion and 1960 Winter Universiade silver medalist.
&
{\bf Article}: Eva Grožajová. ~~~~~~~~~~~~~~~~~~~~~~~~~~~~~~~~~~~~~~~
{\bf Paragraph}: Eva Grožajová , later Bergerová-Grožajová , is a former competitive figure skater who represented Czechoslovakia.  She placed 7th at the 1961 European Championships and 13th at the 1962 World Championships.  She was coached by Hilda Múdra.
&
{\bf Article}: Michael Reusch. ~~~~~~~~~~~~~~~~~~~~~~~~~~~~~~~~~~~~~~
{\bf Paragraph}: Michael Reusch (February 3, 1914–April 6 , 1989) was a Swiss gymnast and Olympic Champion. He competed at the 1936 Summer Olympics in Berlin, where he received silver medals in parallel bars and team combined exercises...
\\\hline
The islands are accessible by a one-hour speedboat journey from Kuala Abai jetty, Kota Belud, 80 km north-east of Kota Kinabalu, the capital of Sabah.
&
{\bf Article}: Mantanani Islands. ~~~~~~~~~~~~~~~~~~~~~~~~~~~~~~~~~~~~~~
{\bf Paragraph}: The Mantanani Islands form a small group of three islands off the north-west coast of the state of Sabah, Malaysia, opposite the town of Kota Belud, in northern Borneo. The largest island is Mantanani Besar; the other two are Mantanani Kecil and Lungisan...
&
{\bf Article}: Gum-Gum ~~~~~~~~~~~~~~~~~~~~~~~~~~~~~~~~~~~~~~~~~~~~~~~~~~~~~
{\bf Paragraph}: Gum-Gum is a township of Sandakan, Sabah, Malaysia. It is situated about 25km from Sandakan town along Labuk Road.
\\\hline
Maggie withholds her conversation with Neil from Tom and goes to the meeting herself, and Neil tells her the spirit that contacted Tom has asked for something and will grow upset if it does not get done.
&
{\bf Article}: Stir of Echoes ~~~~~~~~~~~~~~~~~~~~~~~~~~~~~~~~~~~~~~
{\bf Paragraph}: Stir of Echoes is a 1999 American supernatural horror-thriller released in the United States on September 10 , 1999 , starring Kevin Bacon and directed by David Koepp . The film is loosely based on the novel ''A Stir of Echoes'' by Richard Matheson...
&
{\bf Article}: The Fabulous Five ~~~~~~~~~~~~~~~~~~~~~~~~~~~~~~~~~~~~~~
{\bf Paragraph}: The Fabulous Five is an American book series by Betsy Haynes in the late 1980s .  Written mainly for preteen girls , it is a spin-off of Haynes ' other series about Taffy Sinclair...
\\\hline
\end{tabular}
\caption{StarSpace predictions for some example Wikipedia Article Search (Task 1) queries where StarSpace is correct.
\label{tab:doc-examples}
}
\end{table*}

\subsection{Wikipedia Article Search \& Sentence Matching}
\label{sec:wiki-exp}
In this section, we apply our model on a Wikipedia article search and a sentence match problem. We use the Wikipedia dataset introduced by \cite{chen2017reading}, which is the 2016-12-21 dump of English Wikipedia. For each article, only the plain text is extracted and all structured data sections such as lists and figures are stripped. It contains a total of 5,075,182 articles with 9,008,962 unique uncased token types. The dataset is split into 5,035,182 training examples, 10,000 validation examples and 10,000 test examples. 
\ifarxiv
\else
We then consider the following evaluation tasks:
\fi

\begin{table*}[t!]
\small
\centering
\begin{tabular}{|l|r|r|r|r|r|r|r|r|r|r|r|}
\hline
 Task & MR & CR & SUBJ & MPQA	& SST &	TREC & MRPC & SICK-R & SICK-E & STS14\\
\hline
Unigram-TFIDF* & 73.7 & 79.2 & 90.3 & 82.4 & - & 85.0 & 73.6 / 81.7 & - & - & 0.58 / 0.57 \\
ParagraphVec (DBOW)* & 60.2 & 66.9 & 76.3 & 70.7 & - & 59.4 & 72.9 / 81.1 & - & - & 0.42 / 0.43 \\
SDAE* & 74.6 & 78.0 & 90.8 & 86.9 & - & 78.4 & 73.7 / 80.7 & - & - & 0.37 / 0.38 \\
SIF(GloVe+WR)* & - & - & - & 82.2 & - & - & - & - & 84.6 & 0.69 / - \\
word2vec* & 77.7 & 79.8 & 90.9 & 88.3 & 79.7 & 83.6 & 72.5 / 81.4 & 0.80 & 78.7 & 0.65 / 0.64\\
GloVe* & 78.7 & 78.5 & 91.6 & 87.6 & 79.8 & 83.6 & 72.1 / 80.9 & 0.80 & 78.6 & 0.54 / 0.56 \\
 fastText (public Wikipedia model)* &	76.5 & 78.9 & 91.6 & 87.4 & 78.8 & 81.8 & 72.4 / 81.2 & 0.80 & 77.9 & 0.63 / 0.62 \\	
 StarSpace [word] & 73.8 & 77.5 & 91.53 & 86.6 & 77.2 & 82.2 & 73.1 / 81.8 & 0.79 & 78.8 & 0.65 / 0.62 \\	
 StarSpace [sentence] & 69.1 & 75.1 & 85.4 & 80.5 & 72.0 & 63.0 & 69.2 / 79.7 & 0.76 & 76.2 & 0.70 / 0.67\\
 StarSpace [word + sentence] & 72.1 & 77.1 & 89.6 & 84.1 & 77.5 & 79.0 & 70.2 ／ 80.3 & 0.79 & 77.8 & 0.69/0.66 \\ 
 StarSpace [ensemble w+s] &	76.6 &	80.3 & 91.8 & 88.0 & 79.9 & 85.2 & 71.8 / 80.6 & 0.78 & 82.1 & 0.69 / 0.65 \\
\hline
\end{tabular}
\caption{Transfer test results on SentEval. * indicates model results that have been extracted from \cite{infersent}. For MR, CR, SUBJ, MPQA, SST, TREC, SICK-R we report accuracies; for MRPC, we report accuracy/F1; for SICK-R we report Pearson correlation with relatedness score; for STS we report Pearson/Spearman correlations between the cosine distance of two sentences and human-labeled similarity score.
\label{tab:senteval}
}
\end{table*}

\begin{table*}[t!]
\small
\label{tab:sts}
\centering
\begin{tabular}{|l|r|r|r|r|r|}
\hline
 Task & STS12 & STS13 & STS14 & STS15 & STS16 \\
\hline
fastText (public Wikipedia model) &	0.60 / 0.59 & 0.62 / 0.63 & 0.63 / 0.62 & 0.68 / 0.69 & 0.62 / 0.66\\
StarSpace [word] & 0.53 / 0.54 & 0.60 / 0.60 & 0.65 / 0.62 & 0.68 / 0.67 & 0.64 / 0.65 \\
StarSpace [sentence] & 0.58 / 0.58	& 0.66 / 0.65 & 0.70 / 0.67 & 0.74 / 0.73 & 0.69 / 0.69 \\
StarSpace [word+sentence] & 0.58 / 0.59 & 0.63 / 0.63 & 0.68 / 0.65 & 0.72 / 0.72 & 0.68 / 0.68 \\
StarSpace [ensemble w+s] & 0.58 / 0.59 &  0.64 / 0.64 & 0.69 / 0.65& 0.73 / 0.72 & 0.69 / 0.69 \\
\hline
\end{tabular}
\caption{Transfer test results on STS tasks using
Pearson/Spearman correlations between sentence similarity and human scores. 
. 
\label{tab:senteval2}
}
\end{table*}

\begin{itemize}
\item Task 1: given a sentence from a Wikipedia article as a search query, we try to find the Wikipedia article it came from. We rank the true Wikipedia article (minus the sentence) against 10,000 other Wikipedia articles using ranking evaluation metrics. This mimics a web search like scenario where we would like to search for the most relevant Wikipedia articles (web documents).
Note that we effectively have supervised training data for this task.

\item Task 2: pick two random sentences from a Wikipedia article, use one as the search query, and try to find the other sentence coming from the same original document. We
rank the true sentence against 10,000 other sentences from different Wikipedia articles. This fits the scenario where we want to find sentences that are closely semantically related by topic (but do not necessarily have strong word overlap). Note also that we effectively have supervised training data for this task.

\end{itemize}

We can train our Starspace model in the following way: each update step selects a Wikipedia article from our training set. Then, one random sentence is picked from the article as the input, and for Task 2 another random sentence (different from the input) 
is picked from the article as the label (otherwise the rest of the article for Task 1). Negative entities can be selected at random from the training set. In the case of training for Task 1, for label features we use a feature dropout probability of 0.8 which both regularizes and greatly speeds up training.
We also try StarSpace word-level training, and multi-tasking both sentence and word-level for Task 2.

We compare StarSpace with the publicly released fastText model, as well as a fastText model trained on the text of our dataset.\footnote{FastText training is unsupervised even on our dataset since its original design does not support directly using supervised data here.} We also compare to a TFIDF baseline. For fair comparison, we set the dimension of all embedding models to be 300.
\ifarxiv
\else
The results for tasks 1 and 2 are summarized in Table \ref{tab:wiki1} and \ref{tab:wiki2}  respectively. 
\fi
StarSpace outperforms TFIDF and 
fastText by a significant margin, this is because StarSpace can train directly for the tasks of interest whereas it is not in the declared scope of fastText. Note that StarSpace word-level training, which is similar to fastText in method, obtains similar results to fastText. Crucially, it is StarSpace's ability to do sentence and document level training that brings the performance gains.

A comparison of the predictions of StarSpace and fastText on the article search task (Task 1) on a few random queries are given in Table \ref{tab:doc-examples}. While fastText results are semantically in roughly the right part of the space, they lack finer precision. For example, the first query is looking for articles about an olympic skater, which StarSpace correctly understands whereas fastText picks an olympic gymnast. Note that the query does not specifically mention the word skater, StarSpace can only understand this by understanding related phrases, e.g. the phrase ``Blue Swords'' refers to an international figure skating competition.
The other two examples given yield similar conclusions.


\subsection{Learning Sentence Embeddings}
\label{sec:sentence-exp}
In this section, we evaluate sentence embeddings generated by our model 
and use SentEval\footnote{\tiny\url{https://github.com/facebookresearch/SentEval}} which is a tool from \cite{infersent} for measuring the quality of general purpose sentence embeddings. We use a total of 14 transfer tasks including binary classification, multi-class classification, entailment, paraphrase detection, semantic relatedness and semantic textual similarity from SentEval.
Detailed description of these transfer tasks and baseline models can be found in \cite{infersent}.

We train the following models on the Wikipedia Task 2 from the previous section, and evaluate sentence embeddings generated by those models:
\begin{itemize}
\item StarSpace trained on word level.
\item StarSpace trained on sentence level.
\item StarSpace trained (multi-tasked) on both word and sentence level.
\item Ensemble of StarSpace models trained on both word and sentence level: we train a set of 13 models,
multi-tasking on Wikipedia sentence match and word-level training
 then concatenate all embeddings together to generate a $ 13 \times 300 = 3900 $ dimension embedding for each word. 
\end{itemize}

We present the results in Table \ref{tab:senteval} and 
Table \ref{tab:senteval2}.
StarSpace performs well, outperforming many methods on many of the tasks, although no method wins outright across all tasks. Particularly on the STS (Semantic Textual Similarity) tasks Starspace has very strong results. Please refer to \cite{infersent} for further results and analysis of these datasets. 

\section{Discussion and Conclusion}
\label{sec:conclusion}

In this paper, we propose StarSpace, a method of embedding and ranking entities using the relationships between entities, and show that the method we propose is a general system capable of working on many tasks:

\begin{itemize}
\item Text Classification / Sentiment Analysis: we show that our method achieves good results, comparable to fastText \cite{fasttext} on three different datasets.

\item Content-based Document recommendation: it can directly solve  these tasks well, whereas applying off-the-shelf fastText, Tagspace or word2vec gives inferior results. 

\item Link Prediction in Knowledge Bases: we show that our method outperforms several methods, and matches TransE \cite{transE} on  Freebase 15K.

\item Wikipedia Search and Sentence Matching tasks: it outperforms off-the-shelf embedding models 
due to directly training sentence and document-level embeddings.

\item Learning Sentence Embeddings: It performs well on the 14 SentEval transfer tasks of  \cite{infersent} compared to a host of embedding methods.
\end{itemize}

StarSpace should also be highly applicable to other tasks  we did not evaluate here
such as  other classification, ranking, retrieval or metric learning tasks.
Importantly, what is more general about our method compared to many existing embedding models is: (i) the flexibility of using features to represent labels that we want to classify or rank, which enables it to train directly on a downstream prediction/ranking task;
and (ii) different ways of selecting positives and negatives suitable for those tasks. Choosing the wrong generators $E^+$ and $E^-$ gives greatly inferior results, as shown e.g. in Table \ref{tab:wiki2}.


Future work will consider the following enhancements:
going beyond discrete features, e.g. to continuous features,
considering nonlinear representations and experimenting with other entities such as images. Finally, while our model is relatively efficient, we could consider hierarchical classification schemes as in FastText to try to make it more efficient; the trick here would be to do this while maintaining the generality of our model which is what makes it so appealing.

\section{Acknowledgement}
We would like to thank Timothee Lacroix for sharing with us his implementation of TransE. We also thank Edouard Grave, Armand Joulin and Arthur Szlam for helpful discussions on the StarSpace model.
\bibliographystyle{aaai}
\bibliography{sample}

\begin{thebibliography}{}

\bibitem[\protect\citeauthoryear{Bai \bgroup et al\mbox.\egroup
  }{2009}]{bai2009supervised}
Bai, B.; Weston, J.; Grangier, D.; Collobert, R.; Sadamasa, K.; Qi, Y.;
  Chapelle, O.; and Weinberger, K.
\newblock 2009.
\newblock Supervised semantic indexing.
\newblock In {\em Proceedings of the 18th ACM conference on Information and
  knowledge management},  187--196.
\newblock ACM.

\bibitem[\protect\citeauthoryear{Bengio \bgroup et al\mbox.\egroup
  }{2003}]{bengio2003}
Bengio, Y.; Ducharme, R.; Vincent, P.; and Jauvin, C.
\newblock 2003.
\newblock A neural probabilistic language model.
\newblock {\em Journal of machine learning research} 3(Feb):1137--1155.

\bibitem[\protect\citeauthoryear{Bojanowski \bgroup et al\mbox.\egroup
  }{2017}]{fasttext-unsup}
Bojanowski, P.; Grave, E.; Joulin, A.; and Mikolov, T.
\newblock 2017.
\newblock Enriching word vectors with subword information.
\newblock {\em Transactions of the Association for Computational Linguistics}
  5:135--146.

\bibitem[\protect\citeauthoryear{Bordes \bgroup et al\mbox.\egroup
  }{2011}]{bordes2011learning}
Bordes, A.; Weston, J.; Collobert, R.; Bengio, Y.; et~al.
\newblock 2011.
\newblock Learning structured embeddings of knowledge bases.
\newblock In {\em AAAI}, volume~6, ~6.

\bibitem[\protect\citeauthoryear{Bordes \bgroup et al\mbox.\egroup
  }{2013}]{transE}
Bordes, A.; Usunier, N.; Garcia-Duran, A.; Weston, J.; and Yakhnenko, O.
\newblock 2013.
\newblock Translating embeddings for modeling multi-relational data.
\newblock In {\em Advances in neural information processing systems},
  2787--2795.

\bibitem[\protect\citeauthoryear{Bordes \bgroup et al\mbox.\egroup
  }{2014}]{bordes2014semantic}
Bordes, A.; Glorot, X.; Weston, J.; and Bengio, Y.
\newblock 2014.
\newblock A semantic matching energy function for learning with
  multi-relational data.
\newblock {\em Machine Learning} 94(2):233--259.

\bibitem[\protect\citeauthoryear{Chen \bgroup et al\mbox.\egroup
  }{2017}]{chen2017reading}
Chen, D.; Fisch, A.; Weston, J.; and Bordes, A.
\newblock 2017.
\newblock Reading {Wikipedia} to answer open-domain questions.
\newblock In {\em Association for Computational Linguistics (ACL)}.

\bibitem[\protect\citeauthoryear{Collobert \bgroup et al\mbox.\egroup
  }{2011}]{collobert2011}
Collobert, R.; Weston, J.; Bottou, L.; Karlen, M.; Kavukcuoglu, K.; and Kuksa,
  P.
\newblock 2011.
\newblock Natural language processing (almost) from scratch.
\newblock {\em Journal of Machine Learning Research} 12(Aug):2493--2537.

\bibitem[\protect\citeauthoryear{Conneau \bgroup et al\mbox.\egroup
  }{2016}]{conneau2016very}
Conneau, A.; Schwenk, H.; Barrault, L.; and Lecun, Y.
\newblock 2016.
\newblock Very deep convolutional networks for natural language processing.
\newblock {\em arXiv preprint arXiv:1606.01781}.

\bibitem[\protect\citeauthoryear{Conneau \bgroup et al\mbox.\egroup
  }{2017}]{infersent}
Conneau, A.; Kiela, D.; Schwenk, H.; Barrault, L.; and Bordes, A.
\newblock 2017.
\newblock Supervised learning of universal sentence representations from
  natural language inference data.
\newblock {\em arXiv preprint arXiv:1705.02364}.

\bibitem[\protect\citeauthoryear{Duchi, Hazan, and Singer}{2011}]{adagrad}
Duchi, J.; Hazan, E.; and Singer, Y.
\newblock 2011.
\newblock Adaptive subgradient methods for online learning and stochastic
  optimization.
\newblock {\em Journal of Machine Learning Research} 12(Jul):2121--2159.

\bibitem[\protect\citeauthoryear{Garcia-Duran, Bordes, and
  Usunier}{2015}]{garcia2015composing}
Garcia-Duran, A.; Bordes, A.; and Usunier, N.
\newblock 2015.
\newblock {\em Composing relationships with translations}.
\newblock Ph.D. Dissertation, CNRS, Heudiasyc.

\bibitem[\protect\citeauthoryear{Goldberg \bgroup et al\mbox.\egroup
  }{2001}]{goldberg2001eigentaste}
Goldberg, K.; Roeder, T.; Gupta, D.; and Perkins, C.
\newblock 2001.
\newblock Eigentaste: A constant time collaborative filtering algorithm.
\newblock {\em Information Retrieval} 4(2):133--151.

\bibitem[\protect\citeauthoryear{Hermann \bgroup et al\mbox.\egroup
  }{2014}]{hermann2014}
Hermann, K.~M.; Das, D.; Weston, J.; and Ganchev, K.
\newblock 2014.
\newblock Semantic frame identification with distributed word representations.
\newblock In {\em ACL (1)},  1448--1458.

\bibitem[\protect\citeauthoryear{Jenatton \bgroup et al\mbox.\egroup
  }{2012}]{jenatton2012latent}
Jenatton, R.; Roux, N.~L.; Bordes, A.; and Obozinski, G.~R.
\newblock 2012.
\newblock A latent factor model for highly multi-relational data.
\newblock In {\em Advances in Neural Information Processing Systems},
  3167--3175.

\bibitem[\protect\citeauthoryear{Joulin \bgroup et al\mbox.\egroup
  }{2016}]{fasttext}
Joulin, A.; Grave, E.; Bojanowski, P.; and Mikolov, T.
\newblock 2016.
\newblock Bag of tricks for efficient text classification.
\newblock {\em arXiv preprint arXiv:1607.01759}.

\bibitem[\protect\citeauthoryear{Kadlec, Bajgar, and
  Kleindienst}{2017}]{kadlec2017knowledge}
Kadlec, R.; Bajgar, O.; and Kleindienst, J.
\newblock 2017.
\newblock Knowledge base completion: Baselines strike back.
\newblock {\em arXiv preprint arXiv:1705.10744}.

\bibitem[\protect\citeauthoryear{Koren and Bell}{2015}]{koren2015advances}
Koren, Y., and Bell, R.
\newblock 2015.
\newblock Advances in collaborative filtering.
\newblock In {\em Recommender systems handbook}. Springer.
\newblock  77--118.

\bibitem[\protect\citeauthoryear{Lawrence and Urtasun}{2009}]{lawrence2009non}
Lawrence, N.~D., and Urtasun, R.
\newblock 2009.
\newblock Non-linear matrix factorization with gaussian processes.
\newblock In {\em Proceedings of the 26th Annual International Conference on
  Machine Learning},  601--608.
\newblock ACM.

\bibitem[\protect\citeauthoryear{Lehmann \bgroup et al\mbox.\egroup
  }{2015}]{lehmann2015dbpedia}
Lehmann, J.; Isele, R.; Jakob, M.; Jentzsch, A.; Kontokostas, D.; Mendes,
  P.~N.; Hellmann, S.; Morsey, M.; Van~Kleef, P.; Auer, S.; et~al.
\newblock 2015.
\newblock Dbpedia--a large-scale, multilingual knowledge base extracted from
  wikipedia.
\newblock {\em Semantic Web} 6(2):167--195.

\bibitem[\protect\citeauthoryear{Mikolov \bgroup et al\mbox.\egroup
  }{2013}]{word2vec}
Mikolov, T.; Chen, K.; Corrado, G.; and Dean, J.
\newblock 2013.
\newblock Efficient estimation of word representations in vector space.
\newblock {\em arXiv preprint arXiv:1301.3781}.

\bibitem[\protect\citeauthoryear{Nickel \bgroup et al\mbox.\egroup
  }{2016}]{hole}
Nickel, M.; Rosasco, L.; Poggio, T.~A.; et~al.
\newblock 2016.
\newblock Holographic embeddings of knowledge graphs.

\bibitem[\protect\citeauthoryear{Nickel, Tresp, and Kriegel}{2011}]{rescal}
Nickel, M.; Tresp, V.; and Kriegel, H.-P.
\newblock 2011.
\newblock A three-way model for collective learning on multi-relational data.
\newblock In {\em Proceedings of the 28th international conference on machine
  learning (ICML-11)},  809--816.

\bibitem[\protect\citeauthoryear{Recht \bgroup et al\mbox.\egroup
  }{2011}]{hogwild}
Recht, B.; Re, C.; Wright, S.; and Niu, F.
\newblock 2011.
\newblock Hogwild: A lock-free approach to parallelizing stochastic gradient
  descent.
\newblock In {\em Advances in neural information processing systems},
  693--701.

\bibitem[\protect\citeauthoryear{Rendle}{2010}]{rendle2010factorization}
Rendle, S.
\newblock 2010.
\newblock Factorization machines.
\newblock In {\em Data Mining (ICDM), 2010 IEEE 10th International Conference
  on},  995--1000.
\newblock IEEE.

\bibitem[\protect\citeauthoryear{Shen \bgroup et al\mbox.\egroup
  }{2017}]{shen2017modeling}
Shen, Y.; Huang, P.-S.; Chang, M.-W.; and Gao, J.
\newblock 2017.
\newblock Modeling large-scale structured relationships with shared memory for
  knowledge base completion.
\newblock In {\em Proceedings of the 2nd Workshop on Representation Learning
  for NLP},  57--68.

\bibitem[\protect\citeauthoryear{Shi \bgroup et al\mbox.\egroup
  }{2012}]{shi2012climf}
Shi, Y.; Karatzoglou, A.; Baltrunas, L.; Larson, M.; Oliver, N.; and Hanjalic,
  A.
\newblock 2012.
\newblock Climf: learning to maximize reciprocal rank with collaborative
  less-is-more filtering.
\newblock In {\em Proceedings of the sixth ACM conference on Recommender
  systems},  139--146.
\newblock ACM.

\bibitem[\protect\citeauthoryear{Tang, Qin, and Liu}{2015}]{tang2015document}
Tang, D.; Qin, B.; and Liu, T.
\newblock 2015.
\newblock Document modeling with gated recurrent neural network for sentiment
  classification.
\newblock In {\em EMNLP},  1422--1432.

\bibitem[\protect\citeauthoryear{Weston, Bengio, and Usunier}{2011}]{wsabie}
Weston, J.; Bengio, S.; and Usunier, N.
\newblock 2011.
\newblock Wsabie: Scaling up to large vocabulary image annotation.
\newblock In {\em IJCAI}, volume~11,  2764--2770.

\bibitem[\protect\citeauthoryear{Weston \bgroup et al\mbox.\egroup
  }{2013}]{weston2013connecting}
Weston, J.; Bordes, A.; Yakhnenko, O.; and Usunier, N.
\newblock 2013.
\newblock Connecting language and knowledge bases with embedding models for
  relation extraction.
\newblock {\em arXiv preprint arXiv:1307.7973}.

\bibitem[\protect\citeauthoryear{Weston, Chopra, and Adams}{2014}]{tagspace}
Weston, J.; Chopra, S.; and Adams, K.
\newblock 2014.
\newblock \# tagspace: Semantic embeddings from hashtags.
\newblock In {\em Proceedings of the 2014 Conference on Empirical Methods in
  Natural Language Processing (EMNLP)},  1822--1827.

\bibitem[\protect\citeauthoryear{Xiao and Cho}{2016}]{xiao2016efficient}
Xiao, Y., and Cho, K.
\newblock 2016.
\newblock Efficient character-level document classification by combining
  convolution and recurrent layers.
\newblock {\em arXiv preprint arXiv:1602.00367}.

\bibitem[\protect\citeauthoryear{Zhang and LeCun}{2015}]{zhang2015text}
Zhang, X., and LeCun, Y.
\newblock 2015.
\newblock Text understanding from scratch.
\newblock {\em arXiv preprint arXiv:1502.01710}.

\end{thebibliography}

\end{document}